\documentclass[a4paper,conference]{IEEEtran}
\ifCLASSINFOpdf

\else

\fi

\usepackage{graphicx}
\usepackage{bbding}
\usepackage{amsmath}
\usepackage{graphicx}
\usepackage{mathrsfs}
\usepackage{multirow}
\usepackage{caption}
\usepackage{subfig}
\usepackage{lipsum}
\begin{document}
	\title{ConvMath : A Convolutional Sequence Network for Mathematical Expression Recognition}

	\author{\IEEEauthorblockN{Zuoyu Yan*,
			Xiaode Zhang*,
			Liangcai Gao**,
			Ke Yuan and
			Zhi Tang}
		\IEEEauthorblockA{WICT of Peking University\\ Email: \{yanzuoyu3,zhangxiaode,gaoliangcai,yuanke,tangzhi\}@pku.edu.cn}
	}

	\maketitle

	\begin{abstract}
		Despite the recent advances in optical character recognition (OCR), mathematical expressions still face a great challenge to recognize due to their two-dimensional graphical layout. In this paper, we propose a convolutional sequence modeling network, ConvMath, which converts the mathematical expression description in an image into a LaTeX sequence in an end-to-end way. The network combines an image encoder for feature extraction and a convolutional decoder for sequence generation. Compared with other Long Short Term Memory(LSTM) based encoder-decoder models, ConvMath is entirely based on convolution, thus it is easy to perform parallel computation. Besides, the network adopts multi-layer attention mechanism in the decoder, which allows the model to align output symbols with source feature vectors automatically, and alleviates the problem of lacking coverage while training the model. The performance of ConvMath is evaluated on an open dataset named IM2LATEX-100K, including 103556 samples. The experimental results demonstrate that the proposed network achieves state-of-the-art accuracy and much better efficiency than previous methods.
	\end{abstract}
	
	\IEEEpeerreviewmaketitle

	\section{Introduction}
	\label{intro}
	\let\thefootnote\relax\footnotetext{* Zuoyu Yan and Xiaode Zhang contributed equally to this paper }
	\let\thefootnote\relax\footnotetext{** corresponding author is Liangcai Gao}
	Mathematical expressions are indispensable to describe problems and theories in math, physics, and many other fields, thus play an important role in research and education \cite{zhang2017watch}. Recognition of mathematical expressions aims at converting the two-dimensional (2-D) expression description into a one-dimensional (1-D) structured sequence, which is significant for many downstream applications\cite{zhang2017symbol}, such as scientific document digitization, mathematical information retrieval and reading to visually impaired users. Although some word processing tools support the input of mathematical expressions, it is time-consuming or requires special skills \cite{awal2014global}. For example, LaTeX typing requires a priori knowledge of the markup language grammar and also is slow. Automatic input of the mathematical expression via recognition would bring great convenience to users.
	
	However, due to some characteristics of mathematical expressions, recognition of them is a challenging task. For example, the spatial arrangement of math symbols exhibits complicated 2-D layout, including ABOVE, BELOW, LEFTUP, SUPERscript, SUBscript, CONTAIN, and so on, as mentioned in \cite{zhang2017symbol}. Different from OCR which most commonly recognizes 1-D linear characters one by one, mathematical expression recognition needs to recognize math symbols as well as the 2-D spatial relationships among them, and finally describe the expression using a markup language. In addition, variant scales of math symbols further increase the challenge \cite{zhang2018multi}. Scale variance is influenced by two aspects. One is the symbol position (SUBscript symbols are often smaller than normal symbols) within an expression, the other is size difference caused by input. Besides, a large number of symbols are similar in visual appearances, such as `$a$' and `$\alpha$', `$\Pi$', and `$\prod$', which makes it difficult for the recognizer to distinguish them.
	
	In general, mathematical expression recognition mainly consists of three tasks\cite{zhang2016using}: (1) symbol segmentation, to separate the expression into math symbols; (2) symbol recognition, to assign a class label to each segmented math symbol; (3) structure analysis, to identify the spatial relationships among symbols and generate a 1-D description. These tasks can be solved sequentially or jointly. For sequential cases, the tasks are closely coupled and the errors introduced by the previous step are possible to be inherited to the next step \cite{alvaro2011recognition}. For joint cases, it can process three tasks at the same time, which seems to be a promising solution and has been extensively concerned. However, the joint solution depends on global information, and the complexity of the optimization algorithm increases dramatically with the number of math symbols.
	
	Recently, with the rapid development of deep learning methods, the encoder-decoder model with attention mechanism has been adopted to deal with the mathematical expression recognition problem \cite{zhang2018multi}. The encoder-decoder model has been successfully applied to a variety of tasks such as machine translation \cite{luong2015effective}, image caption \cite{xu2015show}, speech recognition \cite{chan2016listen}, document understanding \cite{onitsuka2019training}, semantic parsing \cite{dong2016language} and question answering \cite{shang2015neural}. This model is commonly employed to transform input data into high-level representations with an encoder, and then the representations are used to generate the target format data by the decoder. With the help of attention mechanism, the decoder can dynamically focus on the most relevant part of the representations, which allows the model to learn soft alignments between input and output \cite{dong2016language}. Mathematical expression recognition is also a suitable application of this model, when the problem is regarded as a transformation from mathematical descriptions (e.g., image) to markup language (e.g., LaTeX). In this way, the model can be trained end-to-end, and the advantage is obvious \cite{zhang2018multi} : (1) Symbol segmentation can be implicitly performed through attention. (2) It is absolutely data-driven and there is no need to define heuristic grammar rules. (3) Scale variance can be handled by the well-designed encoder network.

	In this paper, the encoder-decoder architecture is still adopted but implemented more efficiently. We propose a convolutional sequence modeling network, called ConvMath, to convert the image description of mathematical expressions into LaTeX. The network combines an image encoder adapted from ResNet \cite{he2016deep} for feature extraction and a convolutional decoder modified from \cite{gehring2017convolutional} for sequence generation.

	The remaining part of this paper is organized as follows: Section \ref{sec:1} reviews the related works. Section \ref{sec:arichitecture} introduces the architecture of the proposed convolutional sequence modeling network. The experimental results are presented and discussed in Section \ref{exp}. Finally, the conclusions and future works are presented in Section \ref{Con}.
	
	\section{Related Works}
	\label{sec:1}
	
	Mathematical expression recognition has been an active research area since the late 60s \cite{awal2014global}, and a large number of approaches have been proposed. Survey papers \cite{zanibbi2012recognition} and \cite{chan2000mathematical} have given the detailed summarization. As a supplement, we roughly classified methods for mathematical expression recognition into three categories: rule based, grammar based, and deep learning based \cite{zhang2017symbol}. Previous researches most commonly process the symbol segmentation, symbol recognition, and structure analysis separately, and rule based methods are widely used. Grammar based methods are also prevalent because of their powerful capability to jointly solve these three tasks at the same time. In recent years, the emergence of deep learning provides a new perspective to handle this problem, and the encoder-decoder model has achieved encouraging performance. In the following portion, we mainly present the development of deep learning based methods. 

	Different from rule based methods and grammar based methods, deep learning based methods are absolutely data-driven, which can alleviate the problem caused by manual thresholds, heuristic rules, and predefined grammars. He et al. \cite{he2016context} employ a multi-task network to locate and recognize symbols, which can segment touching and multi-part symbols. However, their network is not able to interpret the expression structure.
	
	From another point of view, the problem of mathematical expression recognition from images can be regarded as a sequence to sequence generation problem, when the image is represented as high-level feature vectors by a convolutional neural network (CNN). Consequently, the methods proposed for sequence-to-sequence are all applicable to this problem.

	Zhang et al. \cite{zhang2016using} adopt the sequence labeling method used in OCR. They represent the mathematical expression with a 1-D sequence derived from the primitive label graph, based on which classical supervised sequence labeling method with connectionist temporal classification (CTC) is applied. This method is not end-to-end trainable, and most researchers prefer the encoder-decoder model as described below.
	\begin{figure*}
		\centering
		
		\includegraphics[width=5.88 in, height=3.116 in]{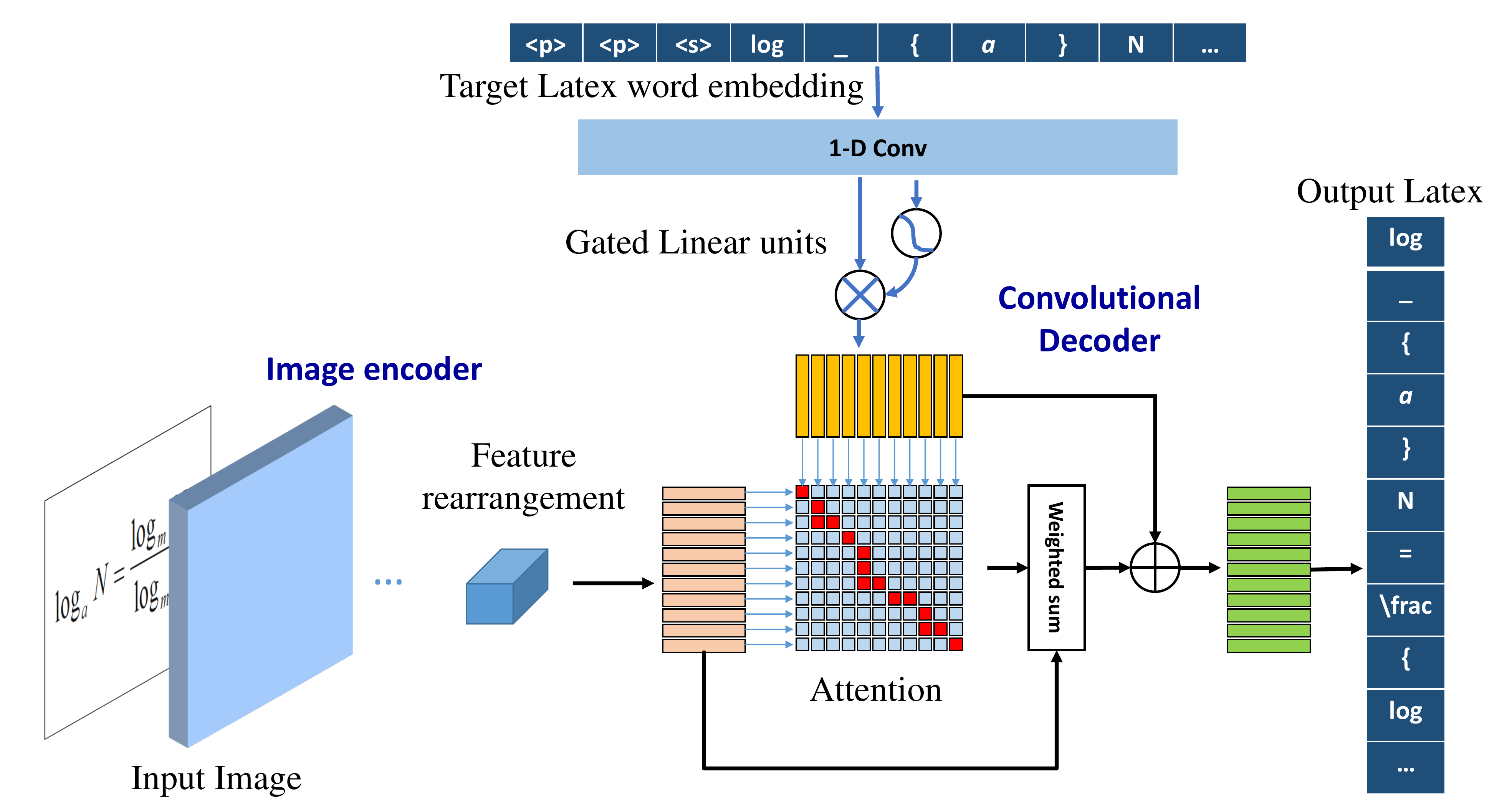}
		\caption{The overall architecture of the proposed model, ConvMath.}
		\label{fig:architecture}       
	\end{figure*}
	
	Encoder-decoder model is prevalent in the field of sequence generation, such as machine translation \cite{luong2015effective}, image caption \cite{xu2015show}, speech recognition \cite{chan2016listen}, document understanding \cite{onitsuka2019training}, semantic parsing \cite{dong2016language} and question answering \cite{shang2015neural}. Deng et al. \cite{deng2016you} present a model, WYGIWYS, which employs a CNN for text and layout recognition in tandem with an attention-based neural machine translation system. It is just a simple extension of the model proposed by Bahdanau \cite{bahdanau2014neural}, and a similar model is also employed in \cite{le2017training} for handwritten mathematical expression. To improve this basic model, Deng et al. introduce a two-layer hard-soft network to attention in \cite{deng2017image}, called scalable coarse-to-fine attention, which can reduce the overhead of attention. Zhang et al. \cite{zhang2017gru} add a coverage-based attention mechanism to the basic encoder-decoder model. By using the alignment history information, this mechanism can effectively alleviate the problem of over parsing or under parsing. In their later work WAP \cite{zhang2017watch}, Zhang et al. integrate a deep fully convolutional neural network (FCN) into the encoder, which enables the model to process large scale input images. To update WAP, a multi-scale attention model \cite{zhang2018multi} is presented to deal with the problems caused by pooling, and DenseNet \cite{huang2017densely} is employed to facilitate feature extraction and gradient propagation. Other improvements such as utilizing residual connection in bidirectional RNN \cite{hong2019residual} and employing both dynamic handwriting traces and static images for feature extraction \cite{wang2019multi} are also impressive.
	
	\section{Network Architecture of ConvMath}
	\label{sec:arichitecture}
	
	The problem of mathematical expression recognition can be described as: given the input image $\boldsymbol{X}$ of width $W$ and height $H$, e.g. $R^{W \times H}$ for gray scale inputs, the goal is to generate the corresponding LaTeX sequence $Y = \{y_1, y_2, \cdots, y_T \}$. $y_{i}$ represents the math symbols in LaTeX string with vocabulary $\boldsymbol{\Omega}$ of size $K$, and $T$ is the length of the sequence. Furthermore, we need to maximize the conditional probability $P(Y|\boldsymbol{X})$ of transforming the image $\boldsymbol{X}$ to LaTeX $Y$ over all dataset $\mathcal{D}$, with respect to the parameters $\theta$ of the model, which can be formulated as:

	\begin{equation}
	\theta ^* = \mathop{\arg\max}_{\theta} \, \sum_{\mathcal{D}} log \, p(Y|\boldsymbol{X})
	\end{equation}
	
	The transformation from image $\boldsymbol{X}$ to LaTeX $Y$ is implemented by the proposed model, ConvMath, as illustrated in Fig.\ref{fig:architecture}, which combines a CNN for feature extraction and a convolutional decoder with attention for LaTeX generation. This model is entirely data-driven and can be trained end-to-end. What's more important, the model can deal with the variable size of the input images, thus there is no need to perform resizing or cropping operation. In the following subsections, we will sequentially introduce the image encoder and convolutional decoder part of the model.
	
	\subsection{Image encoder}
	\label{sec:3.1}
	The high-level visual features of the math images are extracted with a multi-layer CNN. Taken the image of size $W \times H$ as input, the CNN produces an output of size $D \times W^{'} \times H^{'}$, where $D$ denotes the number of channels, $W^{'}$ and $H^{'}$ are the height and width of the feature map. The feature map is a high-level representation of the input image. Each point in the feature map is a $D$ dimensional feature vector and associated with a rectangle region in the original image, which is termed as receptive field. This allows the information in a local area to interact with each other, which is important for the subsequent decision. Intuitively, to determine the 2-D relationships among symbols, the surrounding context information provides significant guidance.
	
	To feed the feature map into the following convolutional decoder, it is rearranged into a sequence of feature vectors $\boldsymbol{V} = \{\boldsymbol{v}_1, \boldsymbol{v}_2, \cdots, \boldsymbol{v}_{W^{'} \times H^{'}} \}$, where $\boldsymbol{v}_{i} \in R^D $. It seems that rearrangement breaks the spatial dependency of the feature vectors, but it doesn't matter because the attention mechanism can dynamically focus on the most relevant part, which is identical to the direct focus on the feature map. Besides, position embedding (described in the following section) further provides the order information of math symbols.
	
	When designing a CNN for feature extraction from mathematical expression images, the following criteria should be taken into consideration: (1) The final feature map requires the combination of both the high-level and low-level representations of the input image. As we all know, the feature representation extracted by CNN becomes more and more abstract with the increase of depth, and the receptive field becomes larger and larger, which is beneficial to modeling 2-D relationships. Meanwhile, because of the pooling operation, deeper features lack detailed information. Math symbols are commonly small objects and detailed information is very useful to identify them. (2) The image encoder should be easy to optimize and keep the capacity at the same time. The whole model, ConvMath, in order to achieve high performance, is relatively sophisticated and deep. The notorious problem of vanishing or exploding gradients easily happens when the model goes deep. Therefore, it is necessary to carefully design the network to ease the flow of gradients.
	
\begin{figure}
		\centering
		\includegraphics[width=3.3 in, height=4.05 in]{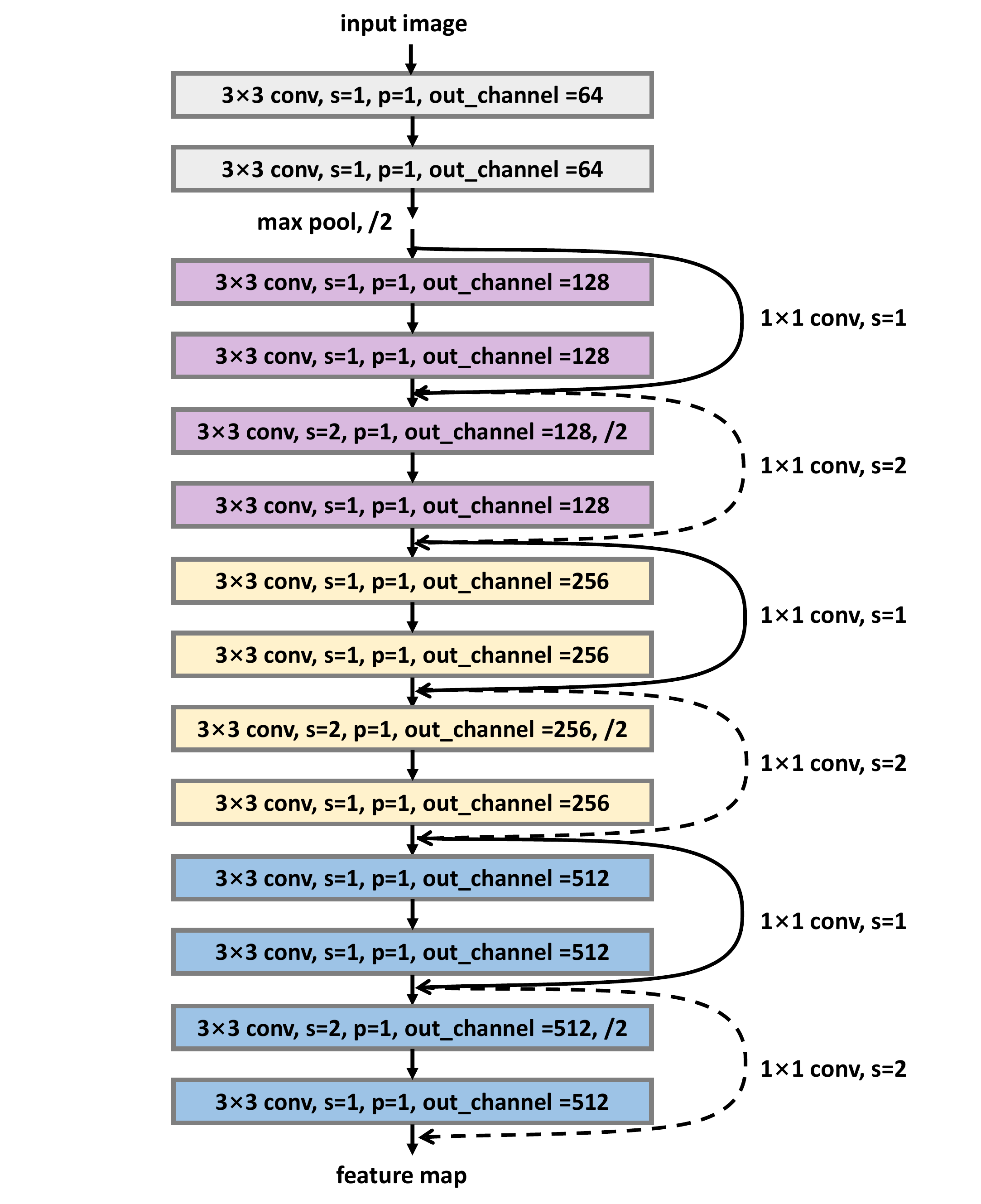}
		\caption{Configurations for image encoder.}
		\label{fig:resnet-encoder}       
	\end{figure}

	Following the above criteria, we modify the residual network (ResNet) \cite{he2016deep} to the proposed image encoder. On the one hand, the residual connection $H(\boldsymbol{x})=F(\boldsymbol{x})+\boldsymbol{x} $ can be viewed as a combination of the low level feature $\boldsymbol{x}$ and high level $F(\boldsymbol{x})$. On the other hand, the shortcut connection simply performs identity mapping, and it allows the fluent gradients propagation. The encoder mainly consists of six residual blocks as shown in Fig. \ref{fig:resnet-encoder}, where $s$ means stride and $p$ means padding. Before the first residual block, two ordinary $3 \times 3$ convolutions are performed on the input image, followed by a $2 \times 2$ max-pooling layer. Each residual block is a stack of 2-layer convolutions with a kernel size of $3 \times 3$. Down-sampling is performed by the convolution layers with the stride of 2, denoted as $/2$ in Fig. \ref{fig:resnet-encoder}. A $1 \times 1$ convolution is employed to match the dimensions (solid line shortcuts) and feature map sizes (dotted line shortcuts) for the residual connections.

	\subsection{Convolutional decoder}
	\label{sec:3.2}
	In this section, we will introduce how to generate the predicted LaTeX sequence $Y = \{y_1, y_2, \cdots, y_{T} \}$ with the convolutional decoder, which takes the feature vectors $\boldsymbol{V} = \{\boldsymbol{v}_1, \boldsymbol{v}_2, \cdots, \boldsymbol{v}_{W^{'} \times H^{'}} \}$ as input, under the supervision of ground truth $W=\{w_1,w_2,\cdots,w_N\}$. This decoder is adapted from the model of Gehring et al. \cite{gehring2017convolutional} developed for sequence-to-sequence learning. Different from other LSTM based encoder-decoder model, this network is entirely convolutional, which means that computations do not depend on the previous time steps and parallelization over every element is available. Besides, variable lengths of input and output are supported, which is vital for the mathematical expression recognition because both the size of image and length of LaTeX string are not fixed.

	\subsubsection{LaTeX embedding and position embedding}
	\label{sec:3.2.1}
	The ground truth of LaTeX $W=\{w_1,w_2,\cdots,w_N\}$ is used to guide the generation of predicted sequence $Y$, where $w_i$ is a LaTeX token. Similar to the word embedding in natural language processing, $W$ is embedded in a distributional space as $\boldsymbol{W} = \{\boldsymbol{w}_1, \boldsymbol{w}_2, \cdots, \boldsymbol{w}_{N} \}$, $\boldsymbol{w}_{i} \in R^D $. To provide the model with order information of math symbols, the absolute position of the symbols is embedded as $\boldsymbol{P} = \{\boldsymbol{p}_1, \boldsymbol{p}_2, \cdots, \boldsymbol{p}_{N} \}$, $\boldsymbol{p}_{i} \in R^D $. These two kinds of embeddings are fused by a simple summation to get the final representation $\boldsymbol{G} = \{\boldsymbol{g}_1, \boldsymbol{g}_2, \cdots, \boldsymbol{g}_{N} \} = \{\boldsymbol{w}_1 + \boldsymbol{p}_1, \boldsymbol{w}_2 + \boldsymbol{p}_2, \cdots, \boldsymbol{w}_N + \boldsymbol{p}_{N} \}$. Position embedding is also applied to the feature vectors $\boldsymbol{V}$. It is important to inform the model which part of the feature maps is currently being processed. For convenience, we still denote the feature vectors after position embedding as $\boldsymbol{V}$.
	
	\subsubsection{Basic block of convolutional decoder}
	\label{sec:3.2.2}
	The convolutional decoder is built by stacking multiple basic blocks, which mainly perform convolution starting from the representation $\boldsymbol{G}$ of ground truth (or generated symbols when testing) layer by layer. To ease the training of network, residual connection between layers is applied. Suppose the output of the $l$-th block is $\boldsymbol{h}^l = \{\boldsymbol{h}_1^l, \boldsymbol{h}_2^l, \cdots, \boldsymbol{h}_{N}^l \}$, the total number of layers is $L$. The residual connection can be formulated as $\boldsymbol{h}^l = conv(\boldsymbol{h}^{l-1})+\boldsymbol{h}^{l-1}$. The prediction of the next math symbol $y_{i+1}$ is based on the previous symbols and feature vectors. Then the conditional probability distribution of $y_{i+1}$ can be formulated as:
	
	\begin{equation}
	p(y_{i+1}|y_1, \cdots, y_i, \boldsymbol{V}) = softmax(\boldsymbol{W}_o\boldsymbol{h}_i^L+\boldsymbol{b}_o) \in R^K
	\end{equation}
	where $\boldsymbol{W}_o$ and $\boldsymbol{b}_o$ are weights and bias of the linear mapping layer, K is the size of vocabulary. Our goal is to minimize the cross entropy loss between the distribution of the prediction and ground truth over all dataset:
	
	\begin{equation}
	L_c = - \frac{1}{|\mathcal{D}|} \sum_{\mathcal{D}}{\sum_{i=1}^{N} log\,p(y_i|y_{<i}, \boldsymbol{X})}
	\end{equation}
	
	The basic block mainly consists of a one dimensional convolution and a subsequent non-linear unit. The 1-D convolution is used to capture the dependencies among LaTeX symbols in the range of kernel width $k$. By stacking more and more basic blocks, larger range information of LaTeX symbols can be obtained. A convolution kernel of width $k$ can be parameterized as $\boldsymbol{W} \in R^{2D \times kD}$, $\boldsymbol{b}_w \in R^{2D}$. It takes $k$ continuous elements in a LaTeX string as input and outputs a 2D-dimensional vector $\boldsymbol{M}\in R^{2D}$. Then, $\boldsymbol{M}$ is split into two D-dimensional vectors, i.e., $\boldsymbol{M} = [\boldsymbol{A}; \boldsymbol{B}]\in R^{2D}$. $\boldsymbol{A} \in R^D$, $\boldsymbol{B} \in R^D$ are used to build non-linear unit (called gated liner units, GLU \cite{gehring2017convolutional}), which transforms $\boldsymbol{M}$ to a $D$ dimensional vector:
	\begin{equation}
	v([\boldsymbol{A}; \boldsymbol{B}]) =\boldsymbol{A} \otimes \sigma(\boldsymbol{B}) \in R^D
	\end{equation}
	where $\otimes$ is the point-wise multiplication and $\sigma(\cdot)$ is sigmoid function, and $v([\boldsymbol{A}; \boldsymbol{B}])$ is then fed into the next block.
	
	The non-linear unit here is very useful to control the information flow between layers. Since convolution treats every element in the receptive field equally, but they should contribute differently to determine the current prediction. Just like the gating mechanism applied in LSTM, GLU can decide which parts of the input should be retained and which part should be discarded, thus can dynamically provide the model with the most important information.
	
	\subsubsection{Attention mechanism}
	\label{sec:3.2.3}

	Attention mechanism can allow the decoder to dynamically focus on the most relevant part of feature vectors when generating symbols. The result of attention is called content vector, which can be viewed as a weighted sum of the feature vectors:
	\begin{equation}
	\boldsymbol{c}_i^l = \sum_{j=1}^{W^{'} \times H^{'}} a_{ij}^l \boldsymbol{v}_j
	\end{equation}
	Here, $ \boldsymbol{c}_i^l $ is the content vector of the $l$-th decoder layer corresponding to the $i$-th state. Once $ \boldsymbol{c}_i^l $ has been computed, it is simply added to the output of decoder layer $\boldsymbol {h}_i^l$ and serves as the input of next layer. $a_{ij}^l$ is the attention score computed as:
	\begin{equation}
	a_{ij}^l = \frac{exp(\boldsymbol{d}_i^l, \boldsymbol{v}_j)}{\sum_{t=1}^{W^{'} \times H^{'}}exp(\boldsymbol{d}_i^l, \boldsymbol{v}_t)}
	\end{equation}
	where
	\begin{equation}
	\boldsymbol{d}_i^l = \boldsymbol{W}_d^l \boldsymbol{h}_i^l+\boldsymbol{b}_d^l+\boldsymbol{g}_i
	\end{equation}
	is the decoder state summary, which combines the current layer output and representation of the previous target element $\boldsymbol{g}_i$. Since $\boldsymbol{h}_i^l$ fuses the information of symbols in the range of receptive field and $\boldsymbol{g}_i$ represents the information of a single symbol. This combination is of great benefit.
	
	It should be noted that the attention mechanism is applied to every decoder layer. This architecture can greatly alleviate the problem of lacking coverage, as mentioned in \cite{zhang2017watch}. Coverage means the overall alignment information that indicates whether a local region of the feature vectors has been translated. Lack of coverage will lead to under or over parsing. Under parsing means some feature vectors are not parsed, while over paring means some symbols are generated multiple times. To address this problem, Zhang et al. \cite{zhang2017watch} append a coverage vector to the attention model, to keep track of past alignment information. In this paper, with the multi-layer attention mechanism, each layer takes the sum of content vector $ \boldsymbol{c}_i^l $ and previous layer output $\boldsymbol{h}_i^l$, i.e. $ \boldsymbol{c}_i^l +\boldsymbol{h}_i^l $, as input. The previous attention information is accumulated and used to calculate the attention score for the next layer. Therefore, we can achieve the tracking of past alignment information without additional coverage vectors.

	\section{Experiments}
	\label{exp}
	To evaluate the performance of the proposed network, we conduct experiments on a large dataset, IM2LATEX-100K, and compare the network with WYGIWYS and WAP. In the following subsections, we will introduce the dataset, evaluation method, and experimental results in detail and the contribution of different parts of the model is also analyzed.
	
	\subsection{Dataset}
	\label{sec:4.1}
	The dataset, IM2LATEX-100K, provided by WYGIWYS \cite{deng2016you}, contains 103,556 different LaTeX expressions extracted from LaTeX sources of over 60,000 papers from the arXiv. The LaTeX source is first converted to PDF files and then converted to PNG format, thus they obtain the images and corresponding ground truth of LaTeX. The LaTeX expression ranges from 38 to 997 characters with mean 118 and median 98. We discard the expressions whose image width is bigger than 480. We follow the same rules as \cite{deng2016you} to tokenize the LaTeX strings. After tokenization, we get a symbol dictionary of size 583. The dataset is randomly separated into the training set (65,995 expressions), validation set (8,181 expressions), and test set (8,301 expressions).
	
\begin{figure}
		\centering
		\includegraphics[width=3.6 in, height=1.98 in]{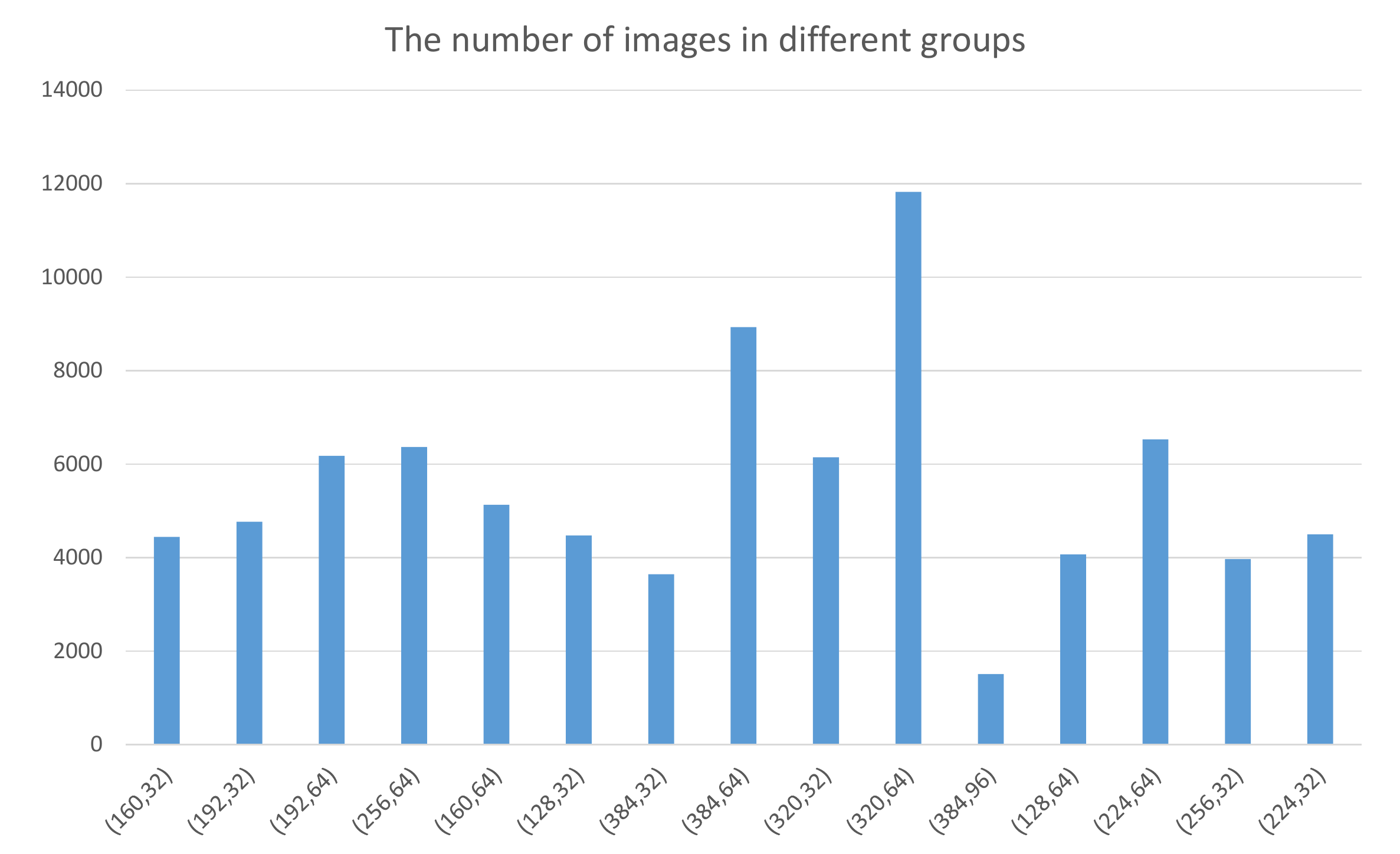}
		\caption{The number of images corresponding to different groups.}
		\label{fig:img_number}       
	\end{figure}

	Although the network has no requirements for the image size of the input, to facilitate batch training, the images are grouped into similar sizes and padded with whitespace while keeping the expression area in the center. There are 15 image groups: (160,32), (192,32), (192,64), (256,64), (160,64), (128,32), (384,32), (384,64), (320,32), (320,64), (384,96), (128,64), (224,64), (256,32), (224,32). The number of images corresponding to different groups is shown in Fig.\ref{fig:img_number}. The LaTeX strings (ground truth) in the same batch are padded to the same length according to the longest length in this batch.

	\subsection{Implementation details}
	\label{sec:4.2}
	
	The configurations for image encoder are presented in Fig. \ref{fig:resnet-encoder}. The embedding size of math tokens is 512, and the vocabulary size is 583. 7 basic blocks (with the kernel size of 3, channel size of 512) are stacked to form the convolutional decoder. The multi-layer attention is implemented with simple dot product, thus there are no parameters to be trained.
	
	The network is implemented with PyTorch 0.4.0 and it is optimized by SGD with an initial learning rate of 0.001, which decays with the rate of 0.8 after 3 epochs. The batch size is set to 15. All the experiments are carried out on a workstation with the Intel(R) Xeon(R) E5-2640 2.50GHz CPU, 64GB RAM, and one NVIDIA Titan X GPU.

	\subsection{Evaluation method}
	\label{sec:4.3}
	Automatic evaluation of the mathematical expression recognition is not an easy task due to the representation ambiguity of the ground truth data (usually as LaTeX or MathML) \cite{zhang2017symbol}. For example, $x'$ can be denoted as \emph{$`x\{ \backslash prime\}$'} or $x'$ in LaTeX. To evaluate the model thoroughly, we adopt these following evaluation methods: (1) the BLEU score. (2) the column-wise edit distance between the original and predicted images. We discretize the generated columns and compute the Levenshtein distance. The final score is the total number of edit distance ops used divided by the maximum number in the dataset. (same as WYGIWYS) (3) the exact match accuracy between the original and predicted images.

	\subsection{Experimental results}
	\label{sec:4.4}
	
	\begin{table}
		\centering
		\caption{Experimental results on IM2LATEX-100K.}
		\label{tab:expres}       
		\scalebox{0.8}{
			\begin{tabular}{lcccc}
				\hline\noalign{\smallskip}
				Method & BLEU & time($s/batch$) & Edit Distance & Exact Match \\
				\noalign{\smallskip}\hline\noalign{\smallskip}
				WYGIWYS\cite{deng2016you} & 87.73 & 0.129 & 87.60 & 79.88 \\
				WAP\cite{zhang2017watch} & 88.21 & 0.135 & 89.58 & 82.08\\
				ConvMath & 88.33 & 0.083 & 90.80 & 83.41\\
				\noalign{\smallskip}\hline
		\end{tabular}}
	\end{table}

	\subsubsection{Comparisons with Other Methods}
	
	The experimental results on IM2LATEX-100K are shown in Table \ref{tab:expres}. The proposed network, ConvMath, can reach the BLEU score of 88.33, which outperforms WYGIWY and WAP. As for running speed, we evaluate the elapsed time to finish a forward inference for a batch data (batch size is 10). WYGIWYS takes 0.129 seconds on average to process a batch, WAP takes 0.135 seconds, while our network only needs 0.083 seconds, which is much faster than WYGIWYS and WAP. It indicates the capability of the convolutional decoder to perform parallel computation, and this structure can extremely speed up the training process.
	
	\subsubsection{Contributions of different parts in the proposed network}
	\label{subsec:different_part}
	
	\begin{table}
		\centering
		\caption{Contributions of different parts in the proposed network.}
		\label{tab:expres_detail}       
		\begin{tabular}{lc}
			\hline\noalign{\smallskip}
			Method & BLEU   \\
			\noalign{\smallskip}\hline\noalign{\smallskip}
			WYGIWYS\cite{deng2016you} & 87.73  \\
			WAP\cite{zhang2017watch} & 88.21\\
			ConvMath\_SimpleEncoder & 80.72  \\
			\noalign{\smallskip}\hline\noalign{\smallskip}
			ConvMath(3 decoder layers) & 84.81  \\
			ConvMath(5 decoder layers) & 87.61  \\
			ConvMath(7 decoder layers) & 88.33  \\
			ConvMath(9 decoder layers) & 88.04  \\
			\noalign{\smallskip}\hline\noalign{\smallskip}
		\end{tabular}
	\end{table}
	
	In this section, we investigate the contributions of different parts in the proposed network. The experimental results of different variants of the network are shown in Table \ref{tab:expres_detail}.
	\begin{table}
		\centering
		\caption{Time evaluation on decoder with ConvMath\_SimpleEncoder}
		\label{tab:time_evaluation}       
		\begin{tabular}{lccc}
			\hline\noalign{\smallskip}
			Decoder & Parameters & time($s/batch$)\\
			\noalign{\smallskip}\hline\noalign{\smallskip}
			CNN decoder(5 decoder layers) & 9313500 & 0.072 \\
			CNN decoder(7 decoder layers) & 13018500 & 0.083\\
			LSTM decoder & 6258500 & 0.129\\
			\noalign{\smallskip}\hline\noalign{\smallskip}
			
		\end{tabular}
	\end{table}\\
	
	\begin{table}
		\centering
		\caption{Evaluation on the residual block of the ResNet encoder}
		\label{tab:res_block}       
		\begin{tabular}{lcc}
			\hline\noalign{\smallskip}
			Encoder & BLEU\\
			\noalign{\smallskip}\hline\noalign{\smallskip}
			ConvMath(4 residual blocks) & 85.37 \\
			ConvMath(6 residual blocks) & 88.33\\
			ConvMath(8 residual blocks) & 87.09\\
			\noalign{\smallskip}\hline\noalign{\smallskip}
			
		\end{tabular}
	\end{table}
	
	\textbf{Contribution of the convolutional decoder:} We justify the contribution of the convolutional decoder based on ConvMath\_SimpleEncoder, which utilizes the same convolutional network as WYGIWYS for feature extraction from images. The configurations for image encoder of WYGIWYS are shown in Table \ref{table1_cong_WYGIWYS}. We evaluate the elapsed time to finish a forward inference for a batch data (batch size is 10) based on different decoders. Table \ref{tab:time_evaluation} shows although the parameter of LSTM decoder is less than that of CNN decoders, CNN decoder with 7 layers are 1.5 times faster than LSTM decoder, which contributes much to long time training. However, the accuracy of CNN decoder with ConvMath\_SimpleEncoder is not so promising, therefore, we introduce residual connection in the encoder to improve the performance of the model.
	
	\textbf{Contribution of the residual encoder:} The modification from ConvMath\_SimpleEncoder to ConvMath(7 decoder layers) is to replace the image encoder with the residual encoder proposed in Section \ref{sec:3.1} This change leads to significant improvement and achieves the best performance. This shows that the features extracted by the residual connection are more expressive than traditional stacked CNN since it combines low-level features and high-level features. 
	
	\textbf{Contribution of the depth of decoder:} The models from ConvMath(3 decoder layers) to ConvMath(9 decoder layers) use the residual image encoder and convolutional decoder but differ in the number of decoder layers (blocks). With the increase of decoder layers, a significant improvement can be observed, but it drops when the number of layers reaches 9. Deep models have larger receptive fields that can capture the relationships among symbols in a larger range. Besides, deep models are also more capable of feature extraction and sequence modeling. However, there is a risk of overfitting when the model goes too deep.
	
	\textbf{Contribution of the depth of encoder:} We modify the number of residual blocks of the encoder to 4 and 8 and change the relative output channel to 256 and 1024, Table \ref{tab:res_block} shows that the performance of the model does not increase significantly while we change the number of residual blocks. To avoid the risk of overfitting and maintain the high speed of the model, we adopt the architecture proposed in the paper.

	\begin{table}
		\centering
		\caption{Configurations for image encoder of WYGIWYS.}
		\label{table1_cong_WYGIWYS}
		\begin{tabular}{l | c | c }
			\hline
			layer name & output size & filter size \\
			\hline
			conv1 & $W$,$H$ & $3 \times 3$, 64  \\

			\hline
			Max pool1 & $W/2$,$H/2$ & $2 \times 2$, 64  \\

			\hline
			conv2 & $W/2$,$H/2$ & $3 \times 3$, 128  \\

			\hline
			Max pool2 & $W/4$,$H/4$ & $2 \times 2$, 128  \\

			\hline
			conv3 & $W/4$,$H/4$ & $3 \times 3$, 256  \\

			\hline
			conv4 & $W/4$,$H/4$ & $3 \times 3$, 256  \\

			\hline
			Max pool3 & $W/8$,$H/4$ & $2 \times 1$, 256  \\

			\hline
			conv5 & $W/8$,$H/4$ & $3 \times 3$, 512  \\

			\hline
			Max pool4 & $W/8$,$H/8$ & $1 \times 2$, 512  \\

			\hline
			conv6 & $W/8$,$H/8$ & $3 \times 3$, 512  \\
			\hline
			
		\end{tabular}
	\end{table}

	\begin{figure}
		\centering
		\includegraphics[width=3.05 in, height=3.57 in]{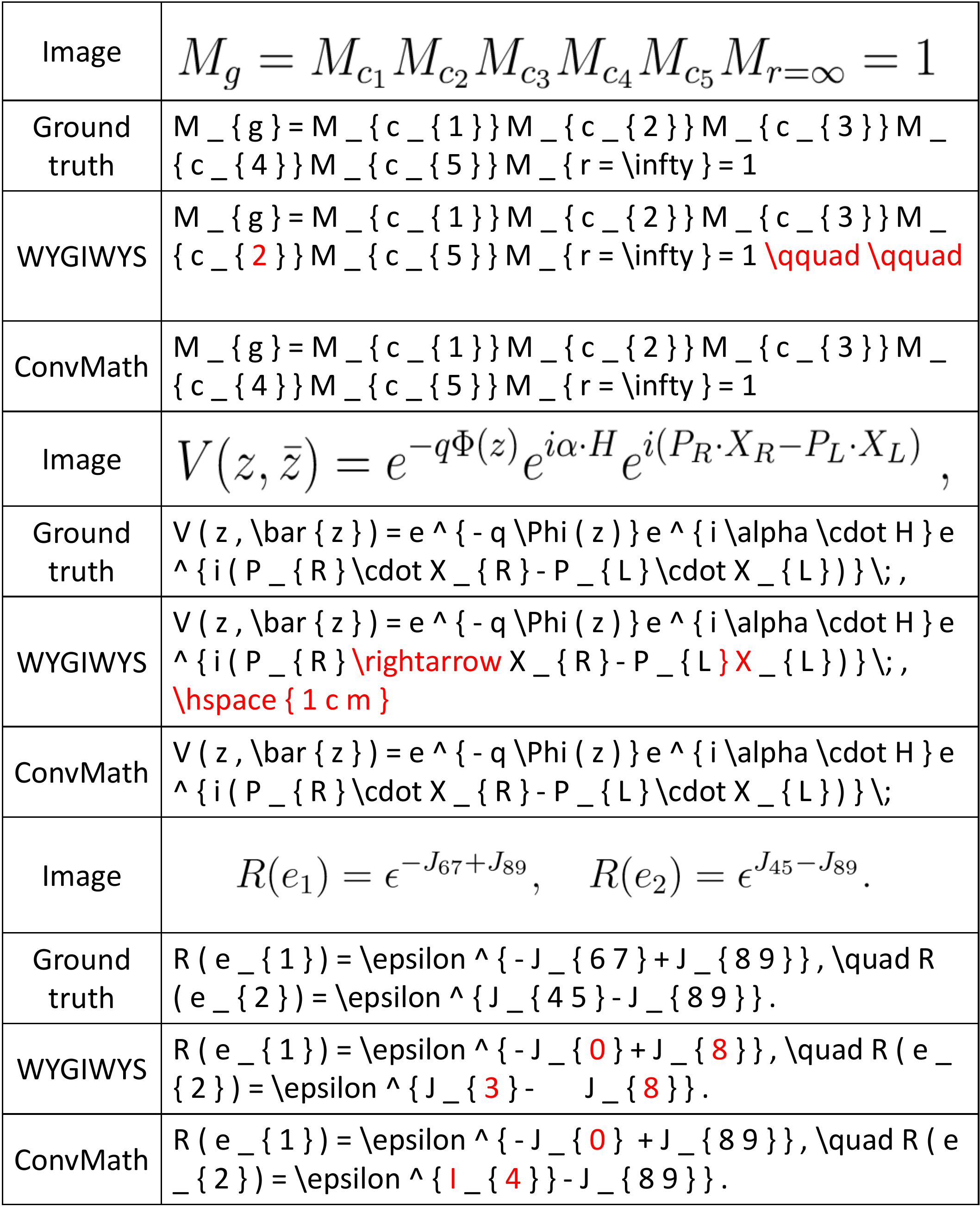}
		\caption{Example of recognition results.}
		\label{fig:case_study}       
	\end{figure}
	
	\subsubsection{Case study}
	
	The effectiveness of multi-layer attention can be verified by the performance improvement from ConvMath\_SimpleEncoder to ConvMath(7 decoder layers). To be more intuitive, some examples of the mathematical expression recognition results are given in Fig.\ref{fig:case_study}. The errors are highlighted in red. We can find that ConvMath achieves higher symbol recognition accuracy. For the first two examples, ConvMath generates completely correct LaTeX strings, while WYGIWYS misrecognizes two symbols. Over parsing rarely happens, but under parsing is common. For the third example, `7', `9', and `5' are discarded by WYYGIWYS. ConvMath performs a little better, only `7' and `9' are missed. Therefore, a future direction is to further strengthen the ability to deal with under parsing problems.

	\section{Conclusion}
	\label{Con}
	In this paper, a neural network is proposed for mathematical expression recognition. The network mainly contains an image encoder for feature extraction and a convolutional decoder for LaTeX generation. With multi-layer attention, the decoder can effectively align the source feature vectors and target math symbols, and the problem of lacking coverage while training the model can be largely alleviated. The model achieves promising recognition results on IM2LATEX-100K, an open mathematical dataset.

	In the future, we will evaluate the network on other datasets like handwritten mathematical expression datasets. We will also explore to apply the network to other tasks such as image caption generation, musical score recognition et al.
	
	\section{ACKNOWLEDGEMENT}
	\label{ack}
	This work is supported by the projects of National Key R\&D Program of China (2019YFB1406303) and National Nature Science Foundation of China (No. 61573028).

	\bibliographystyle{IEEEtran}  
	\bibliography{cite}

\end{document}